\documentclass[10pt,conference]{IEEEtran}

\usepackage{blindtext, graphicx}
\usepackage[labelfont=bf]{caption}
\usepackage{pgfplots}
\usepackage{pgfplotstable}
\pgfplotsset{compat=1.15}
\usepackage{filecontents}
\usepackage{tikz}
\usetikzlibrary{patterns}
\usetikzlibrary{intersections}
\usetikzlibrary{calc}
\usepackage{color}
\usepackage[strings]{underscore}
\usepackage{algorithmicx}
\usepackage[ruled]{algorithm}
\usepackage{algpseudocode}
\usepackage{url}
\usepackage{footnote}
\makesavenoteenv{tabular}
\makesavenoteenv{table}
\usepackage{amsmath,amsthm}
\usepackage{booktabs}
\usepackage{multirow}
\usepackage{comment}
\usepackage{pifont}
\usepackage{pgf-pie}
\usetikzlibrary{shadows}
\usepackage{enumitem}
\usepackage{siunitx}
\usepackage{listings}
\usepackage{parcolumns}
\usepackage{multicol}
\usepackage[utf8]{inputenc}
\usepackage{array}
\DeclareUnicodeCharacter{2212}{-}
\usepackage{colortbl}
\usepackage{tcolorbox}
\usepgfplotslibrary{groupplots}
\usepackage{makecell}
\usepackage[svgpath=./figures/]{svg}
\usepackage{hyperref}
\hypersetup{hidelinks}

\definecolor{dkgreen}{rgb}{0,0.6,0}
\definecolor{gray}{rgb}{0.5,0.5,0.5}
\definecolor{mauve}{rgb}{0.58,0,0.82}
\definecolor{orangep}{rgb}{0.71, 0.43, 0.89}
\definecolor{orp}{rgb}{1, 0.7, 0.278}

\definecolor{darkBlue}{rgb}{0.000000,0.000000,0.545098}
\definecolor{darkGreen}{rgb}{0.000000,0.392157,0.000000}
\definecolor{DarkGray}{gray}{0.4}
\definecolor{javared}{rgb}{0.6,0,0} % for strings
\definecolor{javagreen}{rgb}{0.25,0.5,0.35} % comments
\definecolor{javapurple}{rgb}{0.5,0,0.35} % keywords
\definecolor{javadocblue}{rgb}{0.25,0.35,0.75} % javadoc
\definecolor{lightgray}{gray}{0.95}
\definecolor{shadecolor}{RGB}{150,150,150}
\definecolor{blueA}{RGB}{204,229,255}
\definecolor{redA}{RGB}{112,0, 0}

\lstdefinestyle{MyJavaSmallStyle} {
  language=C++,
  frame=none,
  xleftmargin=15pt,
  stepnumber=1, 
  numbers=left, 
  numbersep=5pt,
  numberstyle=\tiny\color[gray]{0.777}, 
  belowcaptionskip=\bigskipamount,
  captionpos=b, 
  escapeinside={*'}{'*},
  tabsize=5,
  emphstyle={\bf},
  basicstyle=\scriptsize\ttfamily,
  keywordstyle=\color{javapurple}\bfseries,
  stringstyle=\color{javared},
  commentstyle=\color{javagreen},
  morecomment=[s][\color{javadocblue}]{/**}{*/},
  showspaces=false,
  columns=flexible,
  showstringspaces=false,
  morecomment=[l]{//},
  tabsize=2,
  breaklines=true,
  moredelim=[is][\underbar]{^}{^}
}

\lstdefinelanguage{Scala}{
  keywords={typeof, new, true, false, catch,def,val, function, return, null, catch, switch, var, if, in, while, do, else, case, break, assert, static, void ,declare, const, for, define,fun, ite,class, not, check,sat,String, Int, ArrayList},
  keywordstyle=\color{blue}\bfseries,
  ndkeywords={ export,extends, boolean, throw, implements, import, this, abstract,reduceByKey, reduce, filter, map, reduceByKey, join, Join1, public },
  ndkeywordstyle=\color{mauve}\bfseries,
  otherkeywords={+, =>,<=, ==, >,< , ||},
  identifierstyle=\color{black},
  sensitive=false,
  comment=[l]{//},
  morecomment=[s]{/*}{*/},
  commentstyle=\color{purple}\ttfamily,
  stringstyle=\color{red}\ttfamily,
  morestring=[b]',
  morestring=[b]"
}

\algnewcommand\algorithmicforeach{\textbf{for each}}
\algdef{S}[FOR]{ForEach}[1]{\algorithmicforeach\ #1\ \algorithmicdo}

\makeatletter
\newcommand*{\rom}[1]{\expandafter\@slowromancap\romannumeral #1@}

\definecolor{mywhite}{RGB}{255,255,255}
\definecolor{mygray}{RGB}{220,220,220}
\definecolor{olivegreen}{RGB}{0,100,0}

\newif\ifproofread

\proofreadfalse

\usepackage{graphicx} % Required for inserting images
\usepackage[utf8]{inputenc}
\usepackage{float}
\usepackage{subfigure}
\usepackage{pgf-pie}
\usetikzlibrary {datavisualization}
\usepackage{pgfplots}
\pgfplotsset{width=7cm,compat=1.18}
\usepackage{flushend}
\usepackage{amsfonts}

\title{Ultron: Enabling Temporal Geometry Compression of 3D Mesh Sequences using Temporal Correspondence and Mesh Deformation}
\author{Haichao Zhu}
\date{June 2024}

\begin{document}

\maketitle

With the advancement of computer vision, dynamic 3D reconstruction techniques have seen significant progress and found applications in various fields. However, these techniques generate large amounts of 3D data sequences, necessitating efficient storage and transmission methods. Existing 3D model compression methods primarily focus on static models and do not consider inter-frame information, limiting their ability to reduce data size. Temporal mesh compression, which has received less attention, often requires all input meshes to have the same topology, a condition rarely met in real-world applications.

This research proposes a method to compress mesh sequences with arbitrary topology using temporal correspondence and mesh deformation. The method establishes temporal correspondence between consecutive frames, applies a deformation model to transform the mesh from one frame to subsequent frames, and replaces the original meshes with deformed ones if the quality meets a tolerance threshold. 
%For meshes sharing the same topology, motion functions for all vertices’ trajectories are estimated. The motion functions and vertex connectivities are then compressed using entropy-based encoding and corner tables. 
Extensive experiments demonstrate that this method can achieve state-of-the-art performance in terms of compression performance. The contributions of this paper include a geometry and motion-based model for establishing temporal correspondence between meshes, a mesh quality assessment for temporal mesh sequences, an entropy-based encoding and corner table-based method for compressing mesh sequences, and extensive experiments showing the effectiveness of the proposed method. All the code will be open-sourced at \url{https://github.com/lszhuhaichao/ultron}.


\section{Introduction}
In recent years, with the development of computer vision, dynamic 3D reconstruction techniques~\cite{keller2013real,palazzolo2019refusion} for objects such as human body reconstruction~\cite{yu2021function4d} and face reconstruction~\cite{feng2021learning} have made tremendous progress. These dynamic 3D reconstruction techniques are also being applied in many different fields, including 3D games~\cite{metahuman}, virtual reality~\cite{subramanyam2020comparing} and augmented reality~\cite{facepaintingar}. 

However, dynamic 3D reconstruction techniques generate a large amount of 3D data sequences, which further requires more space to store and higher bandwidth to transmit. For example, if a person performs martial arts in a scene and we use a volume capture system to perform 3D reconstruction on it, assuming the FPS (frames per second) is 30, then for a performance that is 1 minute long, we will have 1,800 frames of data, which means we ultimately obtain 1,800 mesh models. When this sequence is transmitted and rendered over the Internet, it is required to complete it within one minute to have a smooth user experience; this places high demands on the transmission and rendering system. 
\begin{figure}
    \centering
    \includegraphics[width=0.5\textwidth]{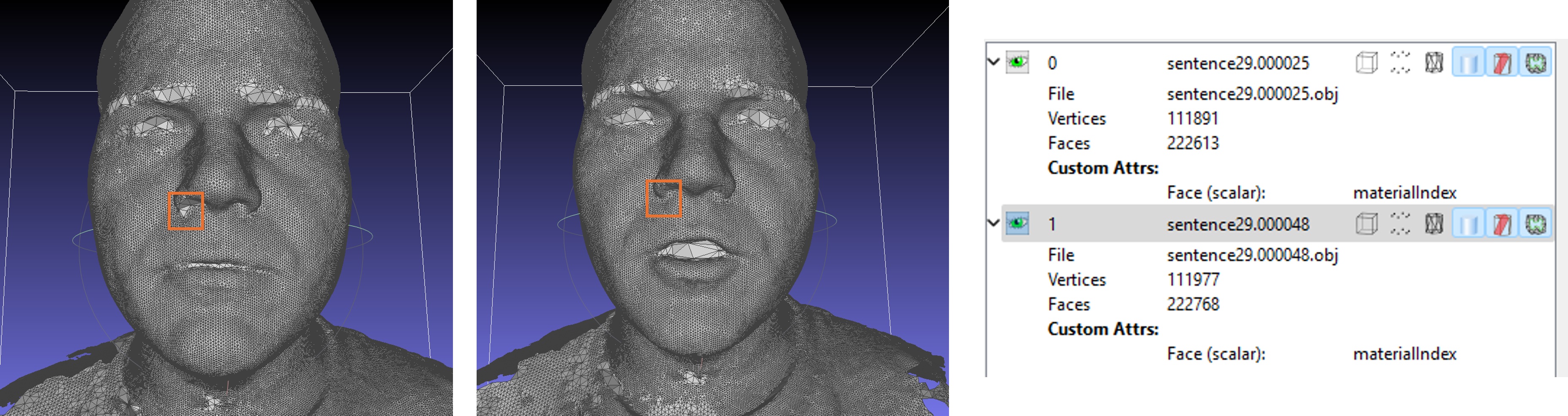}
    \caption{These two frames of meshes are from VOCA~\cite{VOCA2019}. We can see the two mesh are from a same person and look similar; however their topology is different because the number of vertices are different and thus the connectives are also different.}
    \label{fig:meshexample}
\end{figure}

The majority of existing 3D model compression methods primarily focus on static models, meaning they compress each frame of data in a sequence separately, as seen in methods like corner table~\cite{rossignac20013d,draco} or TFan~\cite{mamou2009tfan,Open3DGC}. Consequently, these methods do not consider inter-frame information, limiting their ability to further reduce the data size. An extensive survey on static mesh compression is available in~\cite{maglo20153d}. 
In contrast, temporal mesh compression has received less attention in research. Existing works include MPEG V-DMC~\cite{choi2022overview}, PCA-based methods~\cite{arvanitis2021fast}, and segmentation-based methods~\cite{luo20193d}. However, these methods require that all input meshes have the same topology, a condition that is rarely met in real-world applications due to the limited accuracy of dynamic 3D reconstruction systems. 
An example is provided in Figure~\ref{fig:meshexample}, where two frames of meshes from a sequence are shown. Despite their apparent similarity, these two meshes have different topologies, i.e., they do not have the same number of vertices, and the connectives between vertices are also different. The vertex/edge statistics of these two meshes differ, and the meshes also differ in the two square areas.

In this research, we propose a method to compress mesh sequences with arbitrary topology using temporal correspondence and mesh deformation. Initially, we establish temporal correspondence between consecutive frames based on geometry and motion information. Next, we apply a deformation model to transform the mesh from one frame to subsequent frames. We assess the quality of the deformed meshes compared to their original counterparts using temporal information, and if the quality meets a tolerance threshold, we replace the original meshes with deformed ones while preserving the same topology. 
%For meshes sharing the same topology, we estimate motion functions for all vertices’ trajectories. 
Finally, we compress the vertex coordinates and vertex connectivities using entropy based encoding~\cite{duda2013asymmetric} and corner tables~\cite{rossignac20013d}. Extensive experiments show that our method can achieve state-of-the-art performance in terms of compression performance. In sum, the contribution of this paper can be summarized as follows:
\begin{itemize}
    \item An geometry and motion based model for establishing the temporal correspondence between meshes;
    \item An mesh quality assessment for temporal mesh sequences;
    \item An entropy based encoding and corner table based method for compression mesh sequences.
    \item Extensive experiments show the effectiveness of our proposed method;
\end{itemize}

\section{Related Work}
\subsection{Mesh Matching}
Mesh matching is to find the correspondence between two meshes either in dense or sparse manner. Early methods established such correspondence using manually designed 3D features, like SHOT~\cite{salti2014shot}, PFH~\cite{rusu2008aligning} and FPFH~\cite{rusu2009fast}. \cite{guo2016comprehensive} provides a comprehensive review of such hand-crafted features. Recently, deep learning techniques have been extensive applied to mesh matching problem achieving state-of-the-art performance. Generally, these methods learns new 3D features, like PointNet~\cite{garcia2016pointnet}, 3D oriented histograms~\cite{khoury2017learning} and~\cite{choy2019fully}.

\begin{figure*}[ht!]
    \centering
    \includegraphics[width=\textwidth]{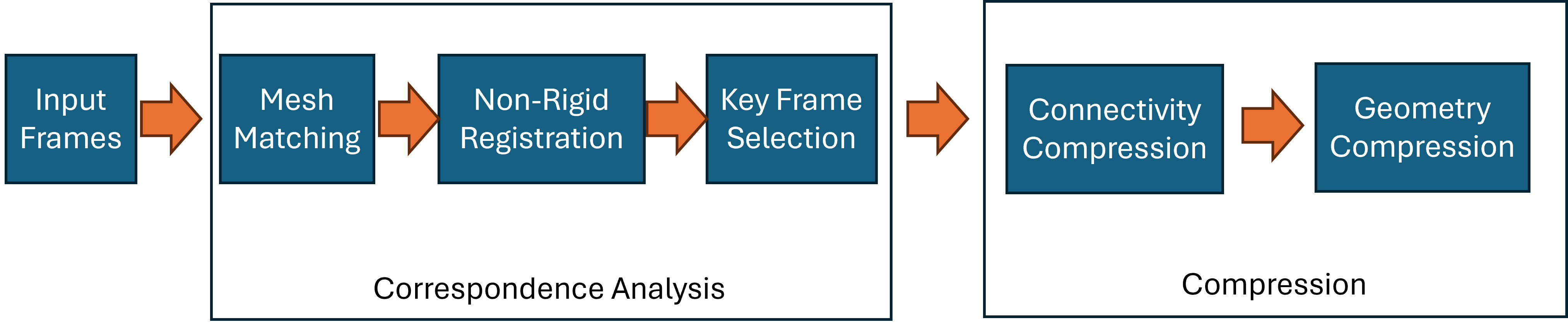}
    \caption{System Overview}
    \label{fig:system}
\end{figure*}

\subsection{Non-rigid 3D registration}
Non-rigid registration is a process that aligns a source surface with a target surface in a flexible way. This implies that different sections of the source surface can experience distinct deformations to accommodate non-rigid behaviors, such as the articulation of the underlying shape.
These methods are generally divided into two categories: extrinsic and intrinsic methods. Extrinsic methods aim to minimize the distance between the source and target surfaces, which is measured in the surrounding 3D space. The alignment can be achieved through an optimization process \cite{amberg2007optimal, yoshiyasu2014conformal, li2020robust} that optimizes an objective function, or by using machine learning techniques to incorporate prior knowledge about the shape, as seen in works like \cite{trappolini2021shape, li2020learning}.
In contrast, intrinsic methods employ intrinsic metrics on the surfaces to calculate the alignment. These intrinsic metrics operate in parametric spaces \cite{sheffer2007mesh}, intrinsic defined distortion \cite{sahilliouglu2018genetic} or spectral domains \cite{hamidian2019surface}.

\subsection{Mesh Compression}
The majority of existing 3D model compression methods fall into two categories: static polygon mesh compression and dynamic mesh sequence compression. Notable static polygon mesh compression methods include corner tables \cite{rossignac20013d,draco} and TFan \cite{mamou2009tfan,Open3DGC}. An extensive survey on static mesh compression is available in \cite{maglo20153d}. However, when these methods are applied to mesh sequences, each frame is usually compressed independently, without utilizing any temporal information. Temporal mesh compression has received less attention in research. Existing works include MPEG V-DMC \cite{choi2022overview}, PCA-based methods \cite{arvanitis2021fast}, and segmentation-based methods \cite{luo20193d}. However, these methods require that all input meshes have the same topology, a condition that is rarely met in real-world applications due to the limited accuracy of dynamic 3D reconstruction systems.

\section{Overview}
The entire system is depicted in Figure~\ref{fig:system}. Our system comprises two components. The first component establishes correspondence between frames, while the second component utilizes this correspondence to compress the mesh sequence. 

In the first component, given a new frame $F_t$ of a 3D mesh, the system initially finds mesh correspondence between $F_i$ and the current key frame $Fk_j$, if the current key frame exists; otherwise, the current frame is set as the current key frame. If the current key frame exists, non-rigid 3D registration is applied to deform the current key frame to the current frame. Subsequently, the quality of the deformed current key frame mesh is evaluated with respect to the current frame. If the quality exceeds a threshold, the deformed current key frame is used to replace the current frame; otherwise, the current frame is set as the current key frame.

In the second component, the mesh sequences are compressed. Frames that are deformed from the same key frame (including the key frame itself) are denoted as a segment. All frames belonging to the same segment are compressed together because their topology is identical, i.e., their connectivity is the same, and this information is compressed only once using existing mesh compression methods like corner table~\cite{draco} or TFAN~\cite{Open3DGC}. For the vertices information, we offer two compression schemes: either use entropy compression to compress the vertices directly or estimate the motion functions of the vertices and compress only the function coefficients.

\section{Method}
In this section, we delve into the specifics of our approach. Initially, we explain how geometry and motion information are utilized to establish vertex correspondence across frames. Following this, we employ a non-rigid 3D registration method using the vertex correspondence to obtain dense vertex correspondence and deform the source mesh to match the target mesh. Subsequently, we introduce a mesh quality evaluation metric that measures both geometric and texture distortions to select key frames. Finally, we discuss how mesh sequences are compressed through connectivity and vertex compression.

\subsection{Mesh tracking}
Consider two consecutive frames at time $t$ and $t+1$, denoted as $F^t$ and $F^{t+1}$, respectively. To establish mesh correspondence, we employ geometry and motion information to match the vertices of the source mesh with those of the target mesh. The vertices of the two meshes are represented as $\mathbf{V}^t=\{v^t_0,v^t_1,...v^t_n\}$ and $\mathbf{V}^{t+1}=\{v^{t+1}_0,v^{t+1}_1,...v^{t+1}_m\}$, where $n$ and $m$ denote the number of vertices of the two meshes, respectively, and $v^t_i, v^{t+1}_j \in \mathbb{R}^3$. Assuming a vertex $v^t_i \in \mathbf{V}^t$ corresponds to a vertex $v^{t+1}_j \in \mathbf{V}^{t+1}$, the motion between them adheres to the second motion function as follows:
\begin{equation}
\begin{aligned}
    v^{t+1}_j &=& v^{t}_i + \int^{t+1}_{t} r_i \mathrm{d}t \\
    r^{t+1}_j &=& r^{t}_i + \int^{t+1}_{t} a_i \mathrm{d}t \\
\end{aligned}
\end{equation}
where $r_i$ and $a_i$ denotes the velocity and acceleration of the vertex $v_i$. We use the the velocity and acceleration at time $t$ to approximate $r_i$ and $a_i$, so the equation become as:

\begin{equation}
\begin{aligned}
    v^{t+1}_j &\approx& v^{t}_i + \int^{t+1}_{t} r^t_i \mathrm{d}t \\
    r^{t+1}_j &\approx& r^{t}_i + \int^{t+1}_{t} a^t_i \mathrm{d}t \\
\end{aligned}
\label{eqn:motionapprox}
\end{equation}
When trying to find the correspondence, we first project each vertex $v^t_i$ from $\mathbf{V}^t$ to a new position $\hat{v}^t_i$ using Equation \ref{eqn:motionapprox}. Then to find the correspondence, we aim to optimize the following the objective function:
\begin{equation}\label{eqn:geometryonj}
\min \sum_i \|\sigma(\hat{v}^t_i)-\sigma(v^{t+1}_j)\|^2
\end{equation}
where $\sigma$ is a geometry function defined on a vertex. We can set $\sigma$ as a  identify function which means the output is just the vertex coordinates or use 3D features, e.g., FPFH~\cite{rusu2009fast}. We can use a dynamic programming to solve Equation~\ref{eqn:geometryonj}.

Typically, our proposed methods allow us to establish vertex correspondence in either a sparse or dense manner. This established correspondence will then serve as the initial setup for the non-rigid 3D registration process.

\subsection{Non-rigid 3D registration}
Consider a key frame $K=\{V^K,E^K\}$ with vertices $V^K=\{v^K_0,v^K_1,...,v^K_m\}$, and a new frame $F=\{V^F,E^F\}$ with vertices $V^F=\{v^F_0,v^F_1,...,v^F_m\}$. Our objective is to apply affine transformations to the vertices $V^K$ of the key frame to align them with the new frame. For each vertex $v^K_i$, we define a $3\times 4$ affine transformation $A_i$. The transformed position $\hat{v}^K_i$ is then given by $A_i\hat{v}^K_i$.

We aim to minimize the distance between the deformed key frame mesh and the current frame mesh. This leads to the first term of the cost function used in non-rigid 3D registration:
\begin{equation}\label{eqn:geometrydist}
    E_{d}(A) = \sum_{\hat{v}^K_i} \textbf{dist}^2(F, A_i\hat{v}^K_i)
\end{equation}
where $\textbf{dist}(F, v)$ is the distance between a point $v$ and its closest point on the frame mesh. 

To regularize the deformation, we introduce a smoothness term that penalizes the weighted difference of the transformations of neighboring vertices:
\begin{equation}
    E_{s}(A) = \sum_{\{i,j\}\in E^F} \|(A_i-A_j)G\|^2_{F}
\end{equation}
where $\|\ddot\|_F$ is the Frobenius norm, and $G=\textbf{diag}(1,1,1,\gamma)$, and $\gamma$ is used to weight differences in the rotational and skew part of the deformation against the translational part of the deformation. 

The third contributor to the cost function is a simple matching term, used for initialization and guidance of the registration which is obtained from the mesh tracking step. Given a set of matching vertices $M=\{(v^K_0,v^F_0),...,(v^K_l ,v^F_l)\}$ mapping key frame vertices into the current frame surface, the matching cost is defined as
\begin{equation}
    E_m(A)=\sum_{(v^K_i,v^F_i)\in M} \|A_iv^k_i-v^F_i\|^2
\end{equation}

Finally, all the cost terms are combined in a weighted sum as the following optimization:
\begin{equation}\label{eqn:finalobjective}
    \min_A E_{d}(A)+\alpha E_{s}(A)+\beta E_{m}(A)
\end{equation}
The smoothness weight $\alpha$ influences the flexibility of the key frame mesh, while the matching weight $\beta$ is used to fade out the importance of the potentially noisy matching towards the end of the registration process.

\subsection{Key frame selection}
We employ two metrics to assess the quality of the deformed mesh, which helps us decide whether or not to insert a new key frame. We use a reference-based scheme to evaluate the quality of the deformed mesh in relation to the original mesh.

The first metric is naturally defined as Equation~\ref{eqn:geometrydist} to quantify the geometric distortion. For the second metric, we utilize an image-based measurement to assess the color difference if the meshes have textures. This definition is based on the $l_2$ norm as follows:
\begin{equation}
E_c = \sum_{{i,j} \in M }|C(v_i)-C(v’_j)|^2_2
\end{equation}
where $M$ represents the set of matching vertices, and $C(\cdot)$ is the function to get the color value of the vertex. If either $E_{d}$ or $E_c$ exceeds a predetermined threshold, the system will insert a new key frame; otherwise the deformed meshes are used to replace the original mesh.

\subsection{Mesh compression}
Once the new mesh sequences are obtained, we can initiate the compression of the meshes. Frames that undergo deformation from the same key frame, including the key frame itself, are classified as a segment. All frames within a given segment are collectively compressed due to their shared topology and, consequently, their identical connectivity. This allows us to compress the connectivity only once. In our system, we can utilize existing mesh compression methods such as the corner table~\cite{draco} or TFAN~\cite{Open3DGC}.

For the compression of vertices with arrtibutes (UV coordinates or vertice normals), we use the same entropy-based compression to compress vertices which are used in~\cite{draco} or TFAN~\cite{Open3DGC}.

\if 0
For the compression of vertices, we employ an adaptive scheme. On one hand, our method involves using entropy-based compression to directly compress vertices, similar to the approach used in~\cite{draco} or TFAN~\cite{Open3DGC}. On the other hand, we estimate the trajectory functions of each vertex and compress only the function coefficients using entropy-based compression. The estimated trajectory is defined as a polynomial function: 
\begin{equation} 
v’_i(t) = a_0+a_1t+a_2t^2 
\end{equation} 
Higher order polynomial functions are also considered during compression. If the estimated trajectory closely aligns with its real trajectory, the adaptive scheme will compress only the coefficients. However, if the movements of the meshes are significant, our scheme will likely opt to directly compress the vertex values.
\fi

\section{Experiments and Results}
\subsection{Dataset and evaluation metrics}
The evaluations are performed on the human motion datasets~\cite{vlasic2008articulated}, and VOCASET from~\cite{VOCA2019} and CTD~\cite{chen2021tightcap}.

\textbf{Human motion datasets}~\cite{vlasic2008articulated} This is 3D human motion dataset. This dataset contains 2 subjects and 10 sequences captured using multi view silhouettes. All the meshes are fitted to the Pinnochio
parametric model~\cite{baran2007automatic}. The meshes do not contain texture information.

\textbf{VOCASET}~\cite{VOCA2019} This is 4D face dataset with about 29 minutes of 4D scans captured at 60 fps and synchronized audio. We only use the scan data fitted with the FLAME parametric model~\cite{FLAME:SiggraphAsia2017}. The dataset has 12 subjects and 480 sequences in total. Since this dataset is for audio-driven animation synthesis, so it does not contains texture information.  

\textbf{CTD}~\cite{chen2021tightcap} We use the dynamic part of CDT. It original has 15 GB, 619 meshes and 14 sequence in total. This data was captured via a dome system using multi view stereo~\cite{schoenberger2016vote} approach or an RGBD-D sensor using DynamicFusion~\cite{newcombe2015dynamicfusion}. Each sequence contains both geometry and texture data. However, the meshes from this dataset may contain irregular structures. So We filter out 4 sequences with data parsing issues and get 4.12 GB, 436 meshes and 10 sequences in total for experiment. 

The summary of the used dataset is listed in Table~\ref{tab:dataset}.

\begin{table*}[t]
    \centering 
    \scalebox{0.9}{
    \begin{tabular}{|c|c|c|c|c|c|c|c|c|}
        \hline
        name & \#Frames & \#Sequences & Parametric & UV & Normal &\#Vertices (Average) & \# Triangles (Average) & Total mesh size (file format) \\
        \hline
        Human motion~\cite{vlasic2008articulated} &2,000 & 12 & Yes & No & No & 10k& 20k & 1.32 GB (.obj)\\
        \hline
        VOCASET~\cite{VOCA2019} & 124K & 480 & Yes & No & No & 5k & 10k & 23.52 GB (.ply binary)\\
        \hline
        CTD~\cite{chen2021tightcap} & 436 &10 & No & Yes & Yes & 30k & 60k & 4.12 GB (.obj) \\
        \hline
    \end{tabular}
    }
     \caption{The summary of data used in our experiments}
    \label{tab:dataset}
\end{table*}

\subsection{Compression performance evaluation}
We first evaluate our method on compression performance. The parameters of the mesh compression is set as following: the quantization bit for the vertex coordinates $qp$, texture UV coordinates $qt$ and normal coordinates $qn$ are set to from 10, 11 and 8 respectively. Note that $qt$ applies when the input meshes have textures and $qn$ applies when the input meshes have vertices normals. No decimation is applied. The compressed file size against original file size are evaluated. Since our contribution is the mesh compression, thus we do evaluate the texture compression performance. The results are given in Table~\ref{tab:resultscompression}. 

\begin{table}[t]
    \centering
    \scalebox{0.9}{
    \begin{tabular}{|c|c|c|c|}
        \hline
        Method & Human motion~\cite{vlasic2008articulated} & VOCASET~\cite{VOCA2019} & CTD~\cite{chen2021tightcap} \\
        \hline
        Original & 1.32 GB & 23.52 GB & 4.12 GB \\
        \hline
        Corner Table (w/o T) & 37.5 MB & 1.11 GB & 113.2 MB \\
        \hline
        Corner Table (with T) & \textbf{31.8 MB} & 965.9 MB & \textbf{42.7 MB} \\
        \hline
        TFAN (w/o T) & 40.6 MB & 1.17 GB & 310.04 MB \\
        \hline
        TFAN (with T) & \textbf{31.8 MB} & \textbf{957.6 MB}  & 397.1 MB \\
        \hline
    \end{tabular}
    }
     \caption{Compression Evaluation}
    \label{tab:resultscompression}
\end{table}

\subsection{Compression quality evaluation}
The quality of the meshes is assessed by comparing the decompressed meshes to the original ones. For meshes without textures, the geometry difference metric is evaluated. For meshes with textures, both geometry and texture information are evaluated. The final results are presented in Table~\ref{tab:resultsquality}. In the Human motion and VOCASET datasets, since the models used are parametric and all vertex connectivity remains the same, there is no compression degradation when using Ultron. However, for the DTC dataset, a degradation in mesh quality is observed when using Ultron.

\begin{table}[t]
    \centering 
    \scalebox{0.9}{
    \begin{tabular}{|c|c|c|c|}
        \hline
        Method & Human motion~\cite{vlasic2008articulated} & VOCASET~\cite{VOCA2019} & CTD~\cite{chen2021tightcap} \\
        \hline
        Corner Table (w/o T) & 67.22 dB & 73.00 dB & 62.82 dB \\
        \hline
        Corner Table (with T) & 67.22 dB & 73.00 dB & 43.08 dB \\
        \hline
        TFAN (w/o T) &67.22 dB & 73.01 dB & 62.83 dB \\
        \hline
        TFAN (with T) &67.22 dB &73.01 dB & 42.14 dB \\
        \hline
    \end{tabular}
    }
       \caption{Mesh Quality Evaluation}
    \label{tab:resultsquality}
\end{table}

\section{Conclusion}
In this paper, we propose a method to compress mesh sequences with arbitrary topology using temporal correspondence and mesh deformation. The method establishes temporal correspondence between consecutive frames, applies a deformation model to transform the mesh from one frame to the next, and replaces the original meshes with deformed ones if the quality meets a tolerance threshold. Extensive experiments demonstrate that this method can achieve state-of-the-art compression performance for parametric models. For non-parametric models, the compression rate is achieved at the cost of mesh quality.
%%For meshes sharing the same topology, motion functions for all vertices’ trajectories are estimated. The motion functions and vertex connectivities are then compressed using entropy-based encoding and corner tables. 

\bibliographystyle{IEEEtran}
\bibliography{papers}
\end{document}